\documentclass[letterpaper, 10pt, conference]{ieeeconf}

\IEEEoverridecommandlockouts                              % This command is only needed if

\usepackage[caption=false]{subfig}
\usepackage{graphicx}
\graphicspath{ {images/} }

\newlength{\myfigwidth}
\newlength{\myfigwidthhalf}
\newlength{\myfigwidththird}
\usepackage[dvipsnames]{xcolor}

\title{\LARGE \bf
\textit{Cartman}: The low-cost Cartesian Manipulator \\that won the Amazon Robotics Challenge
}

\author{%	FirstInitial.~Lastname 	Affiliation		RAS PIN code
D.~Morrison$^{1,2}$,             % 218135
A.W.~Tow$^{1,2}$,                % 191854
M.~McTaggart$^{1,2}$,            % 218492
R.~Smith$^{1,2}$,                % 219035
N.~Kelly-Boxall$^{1,2}$,         % 219057
S.~Wade-McCue$^{1,2}$,\\         % 219110
J.~Erskine$^{1,2}$,              % 219938
R.~Grinover$^{1,2}$,             % 219114
A.~Gurman$^{1,2}$, 				 % 218343
T.~Hunn$^{1,2}$,			     % 219322
D.~Lee$^{1,2}$,                  % 217993
A.~Milan$^{5}$,                % 205393
T.~Pham$^{1,3}$,                  % 192904
G.~Rallos$^{1,2}$,  \\           % 219124
A.~Razjigaev$^{1,2}$,            % 219118
T.~Rowntree$^{1,3}$,             % 218422
K.~Vijay$^{1,3}$,                 % 218987
Z.~Zhuang$^{1,4}$,               % 218333
C.~Lehnert$^{2}$,               % 133477
I.~Reid$^{1,3}$,                % 106056
P.~Corke$^{1,2}$,               % 101638
and J.~Leitner$^{1,2}$% 	    % 154580
\thanks{This research was supported by the Australian Research Council Centre of Excellence for Robotic Vision (ACRV) (project number CE140100016). The participation at the ARC was supported by Amazon Robotics LLC. Contact: {\tt\small douglas.morrison@hdr.qut.edu.au}}% <-this % stops a space
\thanks{$^{1}$Authors are with the Australian Centre for Robotic Vision (ACRV).}%
\thanks{$^{2}$Authors are with the Queensland University of Technology (QUT).}%, Brisbane, QLD 4001 Australia.}%
\thanks{$^{3}$Authors are with the University of Adelaide.}%, SA 5005 Australia.}%
\thanks{$^{4}$ZZ is with the Australian National University (ANU).}%, Canberra, ACT xxxx Australia.}%
\thanks{$^{5}$AM is with Amazon, Germany. This work was done prior to joining Amazon.}
}

\begin{document}

\maketitle
\thispagestyle{empty}
\pagestyle{empty}

\begin{abstract}
The Amazon Robotics Challenge enlisted sixteen teams to each design a pick-and-place robot for autonomous warehousing, addressing development in robotic vision and manipulation. This paper presents the design of our custom-built, cost-effective, Cartesian robot system \textit{Cartman}, which won first place in the competition finals by stowing 14 (out of 16) and picking all 9 items in 27 minutes, scoring a total of 272 points.  We highlight our experience-centred design methodology and key aspects of our system that contributed to our competitiveness. We believe these aspects are crucial to building robust and effective robotic systems.
\end{abstract}

\section{Introduction}

\begin{figure}[t!]
  \centering
    \vspace{1.6mm}
    \includegraphics[trim={1mm 0 0 1mm},clip,width=\myfigwidth]{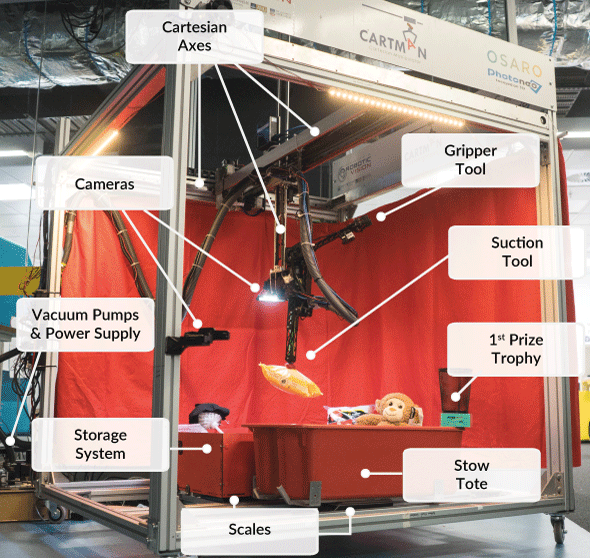}
   \vspace{-4.3mm}
    \caption{\textit{Cartman} %, winner of the Amazon Robotics Challenge,
    is composed of two 6-DoF manipulators, including three shared linear axes, two shared revolute axes, and one exclusive revolute axis each. The wrist holds a camera and two end effector tools, a suction tool and a parallel gripping mechanism. The manipulator is supported by a gantry frame constructed from low cost aluminium extrusions. A secondary camera is attached to the frame, positioned to take a secondary image of picked items with a red backdrop (curtains). The workspace layout is such to minimise travel from the custom storage system to the stow tote and shipping boxes boxes (not pictured).}
  \label{fig:vis-abstract}
  \label{fig:labelledCartman}
  \vspace{-5mm}
\end{figure}

Picking specific items from a cluttered pile of objects is one of the canonical problems in robotics. Applications for robust robotic picking range from household cleaning to space sample return, with considerable economic potential in a wide range of industries including e-commerce and logistics.

Datasets and benchmarks are increasingly exploited in the robotics community to find solutions for such tasks,
yet the applied nature of competitions makes them  one of the greatest drivers of progress -- from self-driving cars, to humanoids, to robotic picking.  The 2017 iteration of the Amazon Robotics Challenge (ARC) (formerly Amazon Picking Challenge) continues this tradition.  The challenge requires autonomous systems to visually detect and identify objects in clutter and then successfully move them in simulated warehouse pick-and-place tasks, requiring robust integration of manipulator design, object recognition, motion planning and robotic grasping.

In this paper we present our robotic vision system, nicknamed \textit{Cartman} (Fig.~\ref{fig:vis-abstract}), which stowed 14 (out of 16) and picked all 9 items in 27 minutes, scoring 272 points to win the challenge finals held in Nagoya, Japan.  The primary differentiating factor of our system is that we use a Cartesian manipulator, which we find greatly simplifies the tasks of planning and executing motions in the confines of the storage system and tote compared to using an articulated robotic arm, and allows us to use a dual-ended end-effector comprising two distinct tools.

In particular, we describe:
\begin{itemize}
    \item The system design of our competition-winning Cartesian robot, including its multi-modal end-effector.
    \item A semantic segmentation perception system capable of learning to recognise novel items from few examples, quickly.
    \item An object-independent grasp synthesis system designed to work under varying levels of visual uncertainty.
    \item Key aspects of our design which contributed to our system's overall performance and robustness.
    \item A set of design principles focused on system integration, based on our Amazon Robotics Challenge experience.
\end{itemize}

\section{Background}

The 2017 Amazon Robotics Challenge comprised two tasks, stow and pick, analogous to warehouse assignments, which involve transferring items between Amazon's totes (a red plastic container), the team's storage system and a selection of Amazon's standard cardboard shipping boxes.  Teams were required to design their own storage system within certain limitations, unlike in previous competitions where standardised shelving units were supplied.  Our storage system comprises two red, wooden boxes with open tops.

In the stow task, teams are required to transfer 20 objects from a cluttered pile within a tote into their storage system within 15 minutes.  Points are awarded based on the system's ability to successfully pick items and correctly report their final location, with penalties for dropping or damaging items, or having items protruding from the storage system.

In the pick task, 32 items were placed by hand into the team's storage system.  The system was provided with an order specifying 10 items to be placed into 3 cardboard boxes within 15 minutes.  Points were awarded for successfully transferring the ordered items into the correct boxes, with the same penalties applied for mishandled or protruding objects.

The 2017 competition introduced a new finals round, in which the top 8 teams competed.  The finals consisted of a combined stow and pick task.  Sixteen items were first hand-placed into the storage system by the team, followed by a vigorous rearrangement by the judges.  Sixteen more items were provided in a tote and had to be stowed into the storage system by the robot.  Then, the system had to perform a pick task of 10 items.  The state of the robot and storage system could not be altered between stow and pick.

A major addition to the challenge compared to previous years was that not all items were known to the teams beforehand.  The items for each task were provided to teams 45 minutes before each competition run, and consisted of 50\% items selected from a set of 40 previously seen items, and 50\% previously unknown items.  This change introduced a major complexity for perception, as systems had to be able to handle a large number of new items in a short period of time.  This, in particular, made deep learning approaches to object recognition much more difficult.

\section{Design Methodology}

\subsection{Experience-centred Design}
Our team, Team ACRV, competed in the 2016 Amazon Picking challenge, placing 6th in the pick task and 14th in the stow task.  Our design methodology for the 2017 competition was heavily driven by our experience and lessons learned, with a major focus on maximising the time spent on system-wide testing rather than testing sub-systems in isolation.

Our general approach to system testing is to perform simulated competition tasks, regularly including contrived scenarios in an attempt to induce system failure.  Through testing, failure modes are recorded and solved where possible.  This practice of adversarial testing on a system level allows us to focus efforts on design aspects which have the largest impact on competition performance, leading to a more robust system overall.

\subsection{Design Practices}

\subsubsection{Modularity}

Our system is designed with software and hardware modularity in mind, making it possible for sub-systems to be developed independently, and easily integrated into the system without requiring changes to higher level systems.  In the case of software, sub-systems largely conform to pre-defined ROS message types throughout development, an example being the perception system which saw major iterations before a final solution was found.  Similarly, changes to the manipulator or Cartesian gantry system can be made and easily integrated into the robot's model, leaving higher level logic unaffected.

\subsubsection{Rapid Iteration}

Iterative design is core to the development of our system.  %During development,
Tasks were broken into weekly sprints, and individual design teams were expected to deliver solutions that could be integrated and tested on the system, a process facilitated by our modular design practices.  This process results in a higher overall integrated system up-time and allowed the team to focus on testing and evaluating the complete system, and to rapidly improve the design at a system or sub-system level as required.

By designing a fully custom solution, we overcame a major disadvantage reported by teams in previous challenges of being locked into the functionality of off-the-shelf components~\cite{correll2016analysis}.  Our design (Fig.~\ref{fig:labelledCartman}) comprises many commonly available parts such as the frame and rails for the Cartesian gantry system, simple machined parts, and a plethora of 3D printed parts.  As such, many aspects of our design are able to be integrated, tested and re-designed within hours or days.

\section{System Design}
\subsection{System Overview}

\begin{figure}[tpb]
  \centering
  \includegraphics[width=2.3in]{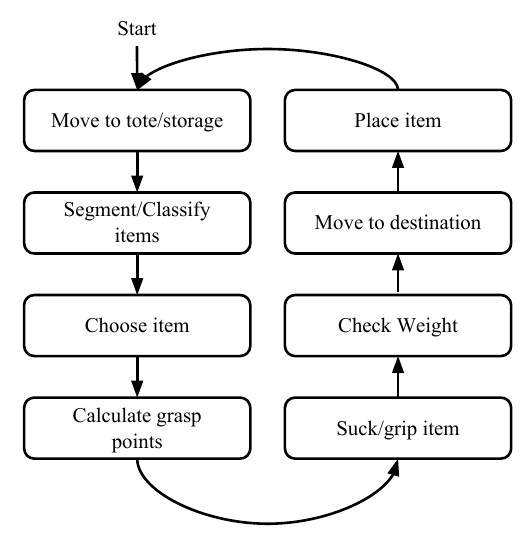}
  \caption{Flow diagram illustrating the main states of execution.}
  \label{fig:statemachine}
  \vspace{-3mm}
\end{figure}

The components of a pick-and-place robot tend to be consistent regardless of the specific application.  At a minimum, they require a perception method for detecting and identifying objects, a method of attaching to objects and a system for manipulating objects.

Fig.~\ref{fig:statemachine} shows an overview of the steps performed by our system in order to perform the stow and pick tasks of the ARC.  \textit{Cartman}'s software is developed on top of the ROS architecture~\cite{quigley2009ros}, which provides a framework for integrating our various sensors, processing subsystems and controllers.  The task-level execution of the system is controlled by a state machine.  In this section we describe the main components of our robotic vision system.

\subsection{Cartesian Manipulator}

The task of picking items from a storage system is mostly a linear problem, which can easily be broken down into straight line motions.  In previous years' challenges, as this one, almost all teams have competed using articulated robotic arms~\cite{correll2016analysis}.  We had previously competed with a Baxter research robot~\cite{ACRVPickingBM}, and encountered difficulties in planning movements of its 7-DoF arm within the confines of an Amazon storage shelf.  Linear movement of an articulated arm requires control of multiple joints, which may result in sections of the arm colliding with the shelf.  In contrast, \textit{Cartman}'s Cartesian design greatly simplifies the task of motion planning within a shelf due to its ability to move linearly along all three axes.  In warehouse pick-and-place tasks, a Cartesian manipulator has many benefits over an articulated robotic arm:
\begin{itemize}
    \item \textit{Workspace}: For pick and place tasks, Cartesian manipulators have a compact rectangular workspace, compared to the circular workspaces of arms.
    \item \textit{Reachability}: Cartesian manipulators allow reaching in a straight line by actuating only a single degree of freedom, resulting in improved reachability within the workspace and lowered chance of collision compared to articulated arms.
    \item \textit{Simplicity}: Motion planning with Cartesian robots is simple,  fast, and unlikely to fail even in proximity to shelving.  On average, motion planning using RRT-Connect~\cite{kuffner2000RRTConnect} planner takes 0.02 seconds.
    \item \textit{Predictability}: The simple, straight-line motions mean that failed planning attempts and erratic behaviour are unlikely.
\end{itemize}

Our system was designed to have specifications comparable to current industrial arms with similar payload capabilities.  The specifications for our system are: a workspace of 1.0m $\times$ 1.0m $\times$ 0.9m; six DoF at each end-effector, given by three linear axes forming the Cartesian gantry and a shared three axis wrist; the capability of actuating with a 2kg payload; a linear velocity of 1 m/s under load along the three linear axes; and an angular velocity of 1 rad/s under load in the angular axes.

For a more in-depth analysis of the hardware design of \textit{Cartman} and its performance  readers are referred to our tech report~%\hl{Cartman Mechanism paper}
\cite{Cartman_mechanism_paper}.

\subsection{Storage System}

A major difference for the 2017 challenge was the requirement for teams to design their own storage system, rather than using a Kiva shelf provided by Amazon.  The main design constraints were a total bounding-box volume of 5000cm$^3$ (roughly the volume of two Amazon totes) and between 2 and 10 internal compartments.  Our design consists of two red, wooden boxes approximately matching the dimensions of the Amazon totes, and opts for a horizontal (top-down picking) design instead of a vertical shelf-like design.  The similarity in colour and size to the totes allows us to use the same perception system for both the tote and the storage system, a design approach taken also by many other teams.

\subsection{Multi-Modal End-Effector}

\begin{figure}[tpb]
  \centering
  \includegraphics[width=\columnwidth]{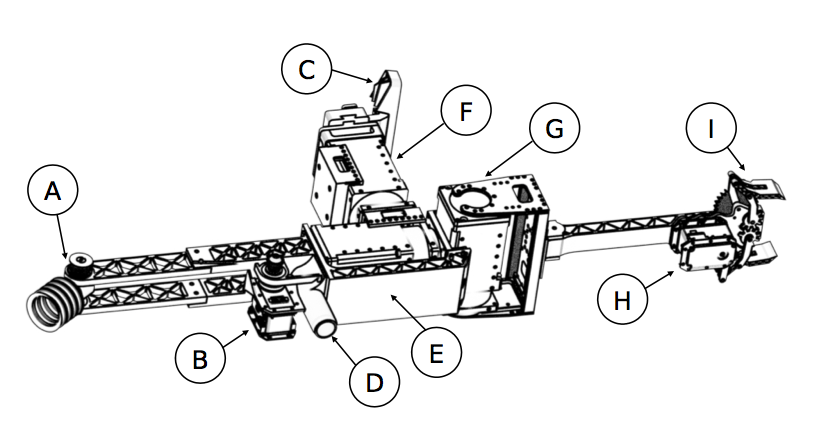}
  \caption{End-effector Assembly. (A) Rotating suction cup (B) Suction gripper pitch servo (drive belt not pictured) (C) wrist-mounted RealSense camera (D) suction hose attachment (E) Roll motor (F) Yaw (tool-change) motor (G) Gripper pitch motor (H) Gripper servo (I) Parallel plate gripper}
  \label{fig:manipulator}
  \vspace{-3mm}
\end{figure}

\begin{figure*}[tb]
    \centering
    \includegraphics[width=\textwidth]{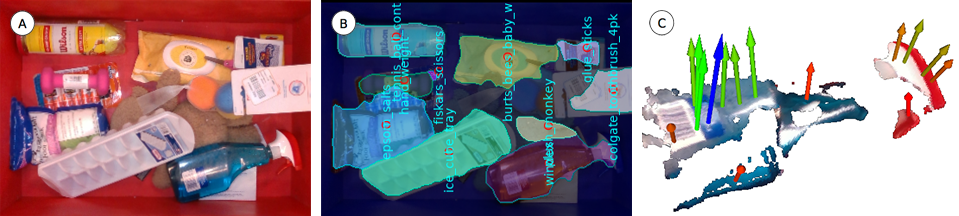}
  \caption{Example of our perception pipeline, showing (A) RGB image of items in our storage system, (B) output of semantic segmentation and (C) segmented point cloud for Windex bottle, with grasp points calculated by the \textit{surface-normals} approach (higher ranked grasp points shown as longer and green and the best candidate in blue).  Even though the semi-transparent Windex bottle results in inaccurate depth information, our system still finds good grasp points on the centre of the bottle, with points on the neck and head being weighted lower.
}
  \label{fig:perception}
  \vspace{-3mm}
\end{figure*}
The ARC requires teams to pick a very diverse set of items, including rigid, semi-rigid, hinged, deformable and porous objects.  Like many other teams, we use a hybrid end-effector design (Fig.~\ref{fig:manipulator}) comprising a suction gripper and a parallel plate gripper.  The suction gripper is our default tool, with the parallel plate gripper being manually specified for porous or deformable objects which don't allow a vacuum seal to be made, e.g. gloves or marbles.

While many other teams combine suction and gripping into a single tool, e.g.~\cite{hernandez2016team, Schwarz2018FastPacking}, the large workspace and Cartesian motion of our robot means that we could easily integrate two distinct tools, selectable by a 180-degree tool change mechanism.  Our design allows each tool to be developed and tested separately, without compromising performance by having to integrate into a single tool.

The suction gripper has a 40mm diameter multi-bellow suction cup, which is able to rotate 180 degrees at its base, allowing it to attach to vertical surfaces inside cluttered environments.  Rotation is belt-driven by a Dynamixel RX-10 servo mounted at the base of the tool to reduce the overall size of the end effector.

Our parallel gripper has a maximum grasping width of 60mm, chosen as the minimum required to suit the subset of items that could not be sucked.  It is powered by a Dynamixel RX-10 servo and has compliant high-friction tape on the gripping faces.

The tool design was based on the set of seen items provided by Amazon.  In an uncluttered environment, our system is capable of picking the 40 seen items with 90\% average reliability over multiple trials.  Worst performance is on the fiskars scissors item, which are small and flat, and the mesh pencil cup, which can not be grasped in all poses.

A more in-depth analysis of the design of our end-effector is provided in \cite{Cartman_endeffector_paper}.

\subsection{Perception Pipeline}

Our perception system provides two key functions: segmentation and identification of items, and generation of grasp points for each (Fig.~\ref{fig:perception}).  For these two components, we use a state-of-the-art semantic segmentation network, RefineNet~\cite{refinenet}, for pixel-wise classification, and a custom vision-based (as opposed to model-based) grasp selection approach. We argue this to be a more robust and scalable solution in the context of picking in cluttered warehouse environments than those based on fitting models or 3D shape primitives, due to issues with semi-rigid, deformable or occluded items.

\subsubsection{Camera Hardware}

Our system uses an Intel RealSense SR300 RGB-D camera, which was chosen primarily due to its small size and low weight which allowed mounting on the robot's wrist (Fig.~\ref{fig:labelledCartman}).  Being on the wrist allows for the camera to be positioned anywhere within the workspace and allowed us to implement the multi-viewpoint active perception approach described in Sec.~\ref{sec:multiview}.  As the RealSense uses an infra-red projector to determine depth, it is unable to produce accurate depth information on some black or reflective items.  We address this issue with the introduction of alternative grasp synthesis techniques for these items (Sec.~\ref{sec:grasppoints}).  A second RealSense camera (Fig.~\ref{fig:labelledCartman}) is mounted on the robot's frame to allow for secondary classification of a picked item if required.

\section{Key System Features}

\subsection{Quick Item Learning}
\label{sec:unknownitems}

A major challenge for the perception system is to learn to recognise the set of unseen items, which are provided only 45 minutes before each competition run, making deep-learning approaches more difficult.  This lead to some other teams using alternative approaches, such as a hybrid feature-matching and convolutional neural network (CNN) approach~\cite{causo2018robust} or metric learning using CNN feature embeddings for singluated objects~\cite{zeng2018robotic}.

After investigating a number of possible approaches, including recent advancements in Deep Metric Learning \cite{G_2016_CVPR}, we found that the best results were obtained by fine-tuning our base RefineNet network on a minimal dataset of the unseen items~\cite{Cartman_vision_paper}.  Our base RefineNet model was initially trained for 200 epochs on a labelled dataset of approximately 200 images of cluttered scenes containing the 40 known items in the Amazon tote or our storage system.

Due to the time-critical nature of learning the unseen items during the competition, we developed a semi-automated data collection procedure which allows us to collect images of each unseen item in 7 unique poses, create a labelled dataset and begin fine-tuning of the network within approximately 7 minutes.  Our procedure is as follows:

\begin{itemize}
    \item Position the wrist-mounted camera above the Amazon tote in the same pose as it would be in during a task.
    \item From an ordered list, place one unseen item in each half of the tote.
    \item Maintaining left/right positioning within the tote, change the orientation and position of each item 7 times and capture an RGB image for each pose.
    \item Using a pre-trained background model, automatically segment each item to generated a labelled dataset.  Each automatically generated training image is manually checked and corrected where necessary.
    \item The new dataset is automatically merged with the existing dataset of known items.
    \item The RefineNet network is fine-tuned on the combined dataset until shortly before the beginning of the official run approximately 30-35 minutes later.
\end{itemize}

During the competition, fine-tuning was performed on an Intel Core i5-7600 and four NVIDIA GTX1080Ti graphics cards.  On this hardware, each training epoch took approximately 130 seconds to compute, usually allowing for 13 epochs to be trained within the time limit.

\begin{figure}[tpb]
  \centering
  \includegraphics[width=\columnwidth]{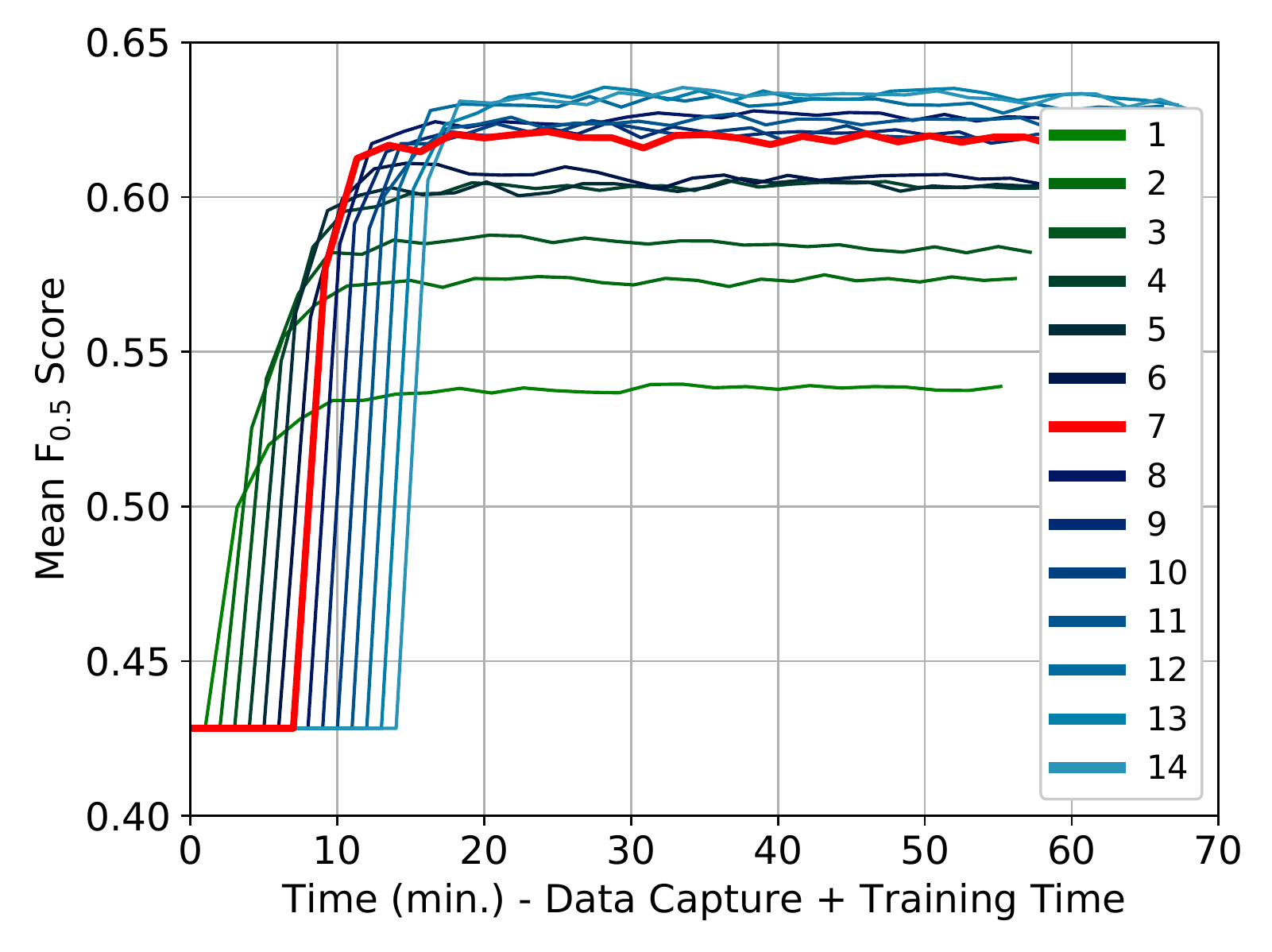}
  \vspace{-4.3mm}
  \caption{Mean F\textsubscript{0.5} score of fine-tuned RefineNet %on a combination of seen and unseen items
  when trained on varying numbers of images for each unseen item.  The time includes data capture and training time.  For the competition, we used 7 images per unseen item, trained for 13 epochs.}
  \vspace{-5mm}

  \label{fig:trainingtime}
\end{figure}

Our data collection procedure is a trade-off between time spent on capturing data and available training time. Fig.~\ref{fig:trainingtime} shows the relative network performance for different splits between number of images captured per unseen item and training time.  For the competition, we opted to use 7 images per unseen item, which allows enough time to repeat the data capture procedure if required.  While it is possible to obtain extra performance by increasing the number of training images, these gains are marginal relative to the amount extra data required.

Second-placed Team NimbRo use a similar perception pipeline, comprising a RefineNet-based architecture which is fine-tuned on synthetic cluttered scenes containing captured images of the unseen items before each competition run~\cite{Schwarz2018FastPacking}.

\begin{figure}[tpb]
  \centering
  \includegraphics[width=\myfigwidth]{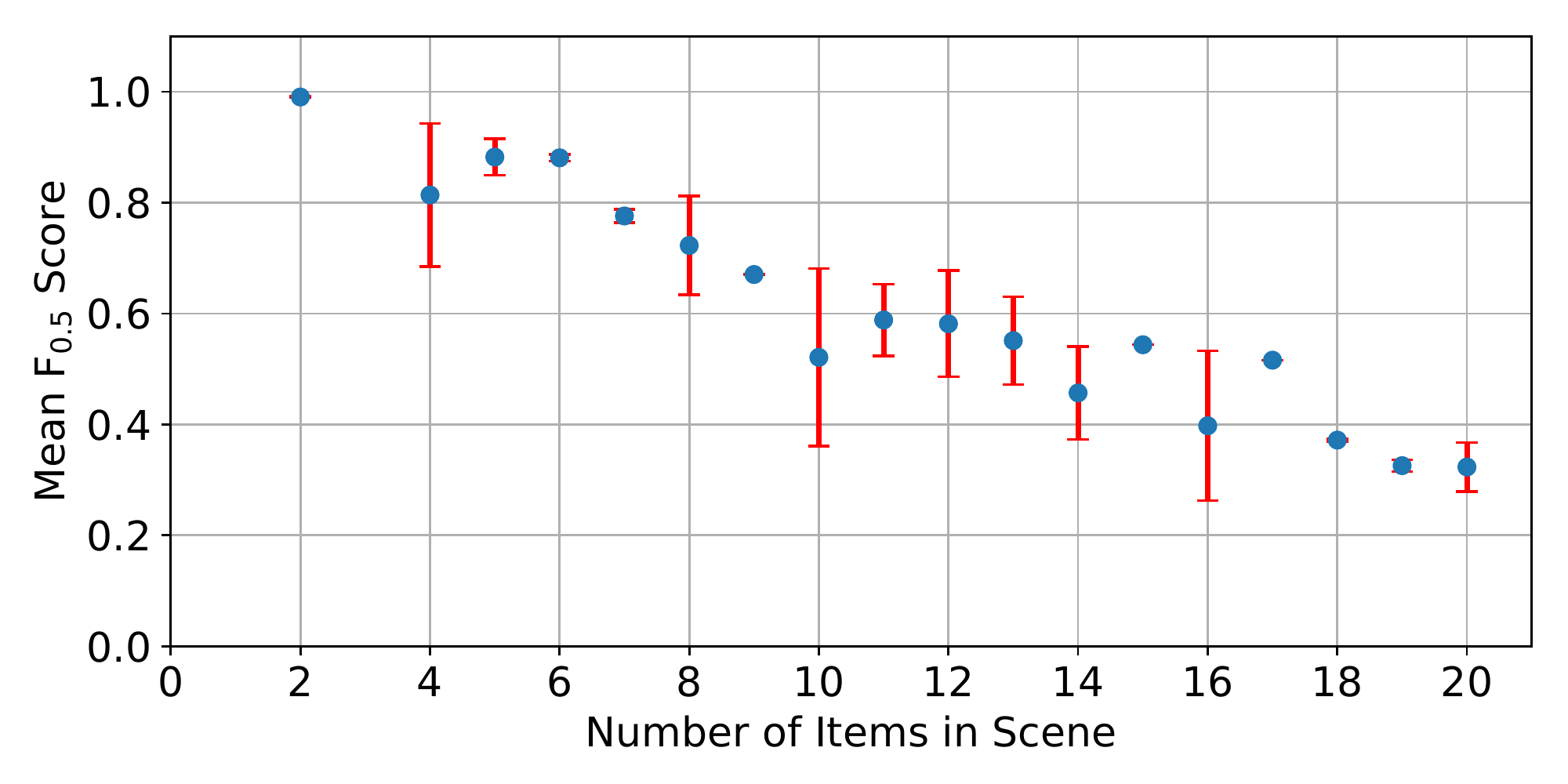}
  \vspace{-4.3mm}
  \caption{Mean F\textsubscript{0.5} score of fine-tuned RefineNet %trained on seen and unseen items
  in increasing levels of clutter.  Error bars denote one standard deviation.}
    \vspace{-5mm}
  \label{fig:refinenetperformance}
\end{figure}

The performance scores shown in Fig.~\ref{fig:trainingtime} are calculated on a set of 67 images representative of those encountered in competition conditions, with a mixture of 1 to 20 seen and unseen items in cluttered scenes.  The network equivalent to that used during the competition (trained on 7 images of each unseen item for 13 epochs) has a mean F\textsubscript{0.5} score of 0.62 on this dataset.  However, Fig.~\ref{fig:refinenetperformance} shows that our network's performance is heavily dependent on the level of clutter in the scene, giving improved performance on scenes containing fewer items.

We use the F\textsubscript{0.5} metric to evaluate the performance of our perception system as it penalises false positives more than false negatives, unlike other commonly used metrics such as IOU or F\textsubscript{1} which penalise false positives and false negatives equally.  We argue that the F\textsubscript{0.5} metric is more applicable for the application of robotic grasping as false positives (predicted labels outside of their item/on other items) are more likely to result in failed grasps than false negatives (labels missing parts of the item).

An in depth analysis of our RefineNet quick-training approach and a comparison with an alternative Deep Metric Learning approach are provided in~\cite{Cartman_vision_paper}.

\subsection{Grasp Synthesis}
\label{sec:grasppoints}

Grasp synthesis is the process of generating grasp parameters for an object by taking into account the physical constraints of the grasping tool, object pose, material properties and surroundings.  For our vision-based grasp synthesis system, the material properties of an item are of particular importance as they effect the quality of the depth information provided by our camera.  To handle all cases, our system uses three different grasp synthesis strategies, which we call \textit{surface-normals}, \textit{centroid} and \textit{RGB-centroid}, which work with varying levels of visual information.  If one method fails to generate valid grasp points for an item, the next method is automatically used at the expense of precision.  For items where the visual information is known to be always unreliable, item meta-data can be provided to the system to always use a specific method.

The next item to pick is chosen to maximise the chances of grasp success.  We group detected items by their relative height in 3cm increments, and select the item in the top-most group with the highest classification certainty.  During this process, items which have failed in the previous three grasp attempts, or do not meet thresholds for segment size or classification certainty are ignored unless no other options are available.  Our grasp synthesis pipeline is then run on the selected item.

The primary approach, \textit{surface-normals}, is used for all items where accurate depth points are available. The process is similar to that described in \cite{Lehnerta}, which calculates grasp points across a segmented point cloud of the object and ranks them using heuristics, illustrated in Fig.~\ref{fig:perception}.  First, a score is calculated for each potential grasp point in a grid on the segmented point cloud, assigning a 75\% weighting to the normalised distance from an object boundary and a 25\% weighting to the object curvature at each point.  We then subtract task-specific penalties of up to 20\% from each grasp score based on its relative height within the storage container and whether the grasp is angled towards the wall of the container.  This method allows grasp trajectories to be generated normal to the surface of the object.

The second approach, \textit{centroid}, is used when objects have valid depth information but that may have been scattered or unreliable, such as with partially reflective objects, resulting in no consistent surfaces on which the \textit{surface-normals} approach can work reliably (i.e. all points are close to a boundary in the point cloud so are rejected).  In this case, a single grasp point is generated at the centre of the object's point cloud, and is executed vertically.

The last approach, \textit{RGB-centroid}, is used when objects have no reliable depth information.  Using the known camera parameters, the position of the object is estimated in 3D space using the centre of the RGB segment.  The grasp is executed vertically, using feedback from the suction sensor and scales to detect when the object has been reached.

Our grasp synthesis system does not require an object pose to generate grasp candidates, but does require a pose estimation to facilitate anti-podal gripping and aligning items for packing.  To estimate the pose of the item to be grasped, we compute the PCA of the item's RGB segment and align the gripper or sucker to its principal orientation.  We find that this approach performs well except in cases of major occlusion.

One aspect of the grasping system which sets \textit{Cartman} apart from other participants of the ARC is the ability to execute multiple grasp attempts sequentially.  For suction grasps, the 3 best, spatially diverse grasp points are stored, and if suction fails on one the next is automatically tried without having to perform another visual processing step, increasing the chances of succeeding in the grasp and increasing overall system speed.

\subsection{Active Perception}
\label{sec:activeperception}

For the stow task, where all items have to be picked, we preference picking unoccluded items at the top of the tote, which counteracts the tendency for our perception network to exhibit lower precision in more cluttered scenes (Fig.~\ref{fig:refinenetperformance}).  However, for the pick task, where only a subset of items have to be picked, it is likely that not all wanted items are easily visible within the storage system.  In this task, we use two active perception methods to increase the chances of finding all of the wanted items.

\subsubsection{Multi-view}
\label{sec:multiview}

For each storage compartment, there are three possible camera poses; one top view capturing the entire storage compartment, and two close-up views covering half of the storage compartment each.  If no wanted objects are visible from the top view, the two close-up views are used, leveraging the adjusted camera viewpoint to increase the chances of finding any partially obscured items, and reducing the effective level of clutter thereby improving the performance of our perception system on the scene.

\subsubsection{Directed Item Search}

In the case that no wanted items are visible from any viewpoint, our system will attempt to find them by moving unwanted items, either within a single storage compartment, or between compartments if all wanted items have been removed from the other.  Two methods are used to increase the chances of finding wanted objects quickly.  First, a history of previous classification outputs is used to determine if any wanted items were previously visible.  If so, items that were observed near the wanted items are moved first.  Second, if no wanted items have been previously observed, items are chosen based on the likelihood that they are obscuring items by taking into account their size and relative height within the storage system.

In addition to potentially exposing obscured items, the case of moving items between storage compartments has the effect of improving the precision of our perception network by reducing the number of items in the storage compartment which contains the wanted item(s).

\subsection{System Robustness}
Our system level testing lead to a large focus on robustness and error recovery in developing high-level logic for \textit{Cartman}.  In this section we discuss three key areas of our design which contributed most significantly to our system's overall performance, and to our win in the competitions finals.

\subsubsection{Extra Sensing}
In addition to the primary camera, our system also relies on a number of other sensors for object recognition and error detection.  The tote and storage system are all supported on scales, which are used to confirm that the correct item is grasped, that it wasn't dropped and to detect when the end-effector has come into contact with an item.  A pressure switch is included on the vacuum line to detect when a suction seal is made, and to detect dropped items.  A second camera is mounted on the outside of the robot frame, facing across the workspace, to allow for a second visual classification on a grasped object if required.

\subsubsection{Item Reclassification}

To be sure of a grasped object's identity, our system requires consensus from two sensors.  The first is by the primary visual classification.  Secondly, when a grasped item is lifted, the weight different measured by the scales is used to confirm the object's identity.  If the measured weight does not match, one of two reclassification methods are used.  If the measured weight matches only one item in the source container then the item is immediately reclassified based on weight and the task continues uninterrupted.  Alternatively, if there are multiple item candidates based on weight, the item is held in view of the side-mounted camera to perform a second visual classification of the item.  If the item is successfully classified as one of the candidates, it is reclassified and the task continues.  If no suitable classification is given, the item is replaced and the next attempt begins.

\subsubsection{Error Detection and Recovery}
Like with any autonomous system designed to operate in the real world, things always can and eventually will go wrong.  As such, the robot's sensors are monitored throughout the task to detect and automatically correct for failures, such as failed grasps, dropped items and incorrectly classified or re-classified items.  During the long term testing described in Sec.~\ref{sec:systemperformance}, only 4\% (10 out of 241) of failures were not corrected, requiring manual intervention for the robot to finish a task.  These failures were caused by grasping an incorrect item that happened to have the same weight as the wanted item, picking two items together and reclassifying by weight as a different item, or dropping the item outside the workspace.

\textit{Cartman}'s actions are heavily dependent on having an accurate internal state of the task, including the locations of all items.  As an extra layer of redundancy, towards the end of a task when there are 5 or fewer items remaining, the system begins to visually double-check the location of items.  If an item is consistently detected in the wrong location, the system attempts to correct its internal state based on visual classification and a record of any previous item reclassifications.  This is possible due to the false negative rate of our classifier being much lower in uncluttered scenes, such as at the end of a run (Fig.~\ref{fig:refinenetperformance}).

\section{System Performance}
\label{sec:systemperformance}

\begin{table}[tpb]
    \caption{Speed and accuracy of all systems during finals task and our long-term test (ACRV-LT).}
    \label{tab:speed_results}
    \begin{tabular}{|c||c|c|c||c|}
        \hline
        & \textbf{Grasp} & \textbf{Avg.} & \textbf{Error} & \textbf{Final}  \\
         \textbf{Team} & \textbf{Success Rate} & \textbf{Time} & \textbf{Rate} & \textbf{Score }\\
        \hline
        \hline
        Applied Robotics & 50\% (3/6) & 101s & 0\% (0/2) & 20 \\
        \hline
        IFL PiRo & 78\% (18/23) & 59s & 50\% (7/14) & 30 \\
        \hline
        NAIST-Panasonic & 49\% (21/43) & 35s & 33\% (5/15) & 90 \\
        \hline
        MIT-Princeton & 66\% (43/65) & 25s & 0\% (0/15) & 115 \\
        \hline
        IITK-TCS & 79\% (19/24) & 40s & 15\% (3/20) & 170 \\
        \hline
        Nanyang & 53\% (23/43) & 32s & 4\% (1/25) & 225 \\
        \hline
        NimbRo Picking & 58\% (33/57) & 29s & 0\% (0/22) & 235 \\
        \hline
        \textbf{ACRV} & 63\% (33/52) & 30s & 4\% (1/23) & 272 \\
        \hline
        \hline
        \textbf{ACRV-LT} & 72\% (622/863) & 30s & N/A & N/A \\
        \hline
        \end{tabular}
\vspace{-3mm}
\end{table}

To test the overall performance of the system, \textit{Cartman} was run for a full day, performing a continuous finals-style task, where the pick phase was used to replace all items from the storage system into the tote.  17 items were used, consisting of 13 items from the Amazon item set and 4 unseen items from the ACRV picking benchmark set~\cite{ACRVPickingBM}.  The 4 unseen items were a soft, stuffed animal \texttt{plush\_monkey}, a reflective, metallic pet's food bowl \texttt{pets\_bowl}, a scrubbing brush \texttt{utility\_brush} and colourful squeaky pets toys \texttt{squeaky\_balls}, which were chosen for their similarity to unseen items provided by Amazon during the competition.  The 17 items were chosen to provide a range of difficulties as well as cover the spectrum of object classes that were available both physically (rigid, semi-rigid, deformable and hinged) and visually (opaque, partially transparent, transparent, reflective and IR-absorbing), ensuring that the full range of \textit{Cartman}'s perception and grasping abilities were tested.  Nine of the objects were acquired using suction and  8 by the gripper.  Arguably the hardest item in the Amazon set, the mesh cup, was not included as \textit{Cartman} was unable to grasp this item when it was on its side.

In 7.2 hours of running time, \textit{Cartman} completed 19 stow and 18 pick tasks, during which 863 grasping attempts were performed, 622 of which were successful (72\% success rate, ACRV-LT in Table~\ref{tab:speed_results}).  Throughout the experiment, 10 items were incorrectly classified without automatic correction, requiring manual intervention to allow the system to complete a task and continue with the next.  On one occasion the system had to be reset to correct a skipped drive belt.

\begin{figure}[tpb]
  \centering
  \includegraphics[trim={0 5mm 0 0},clip,width=\columnwidth]{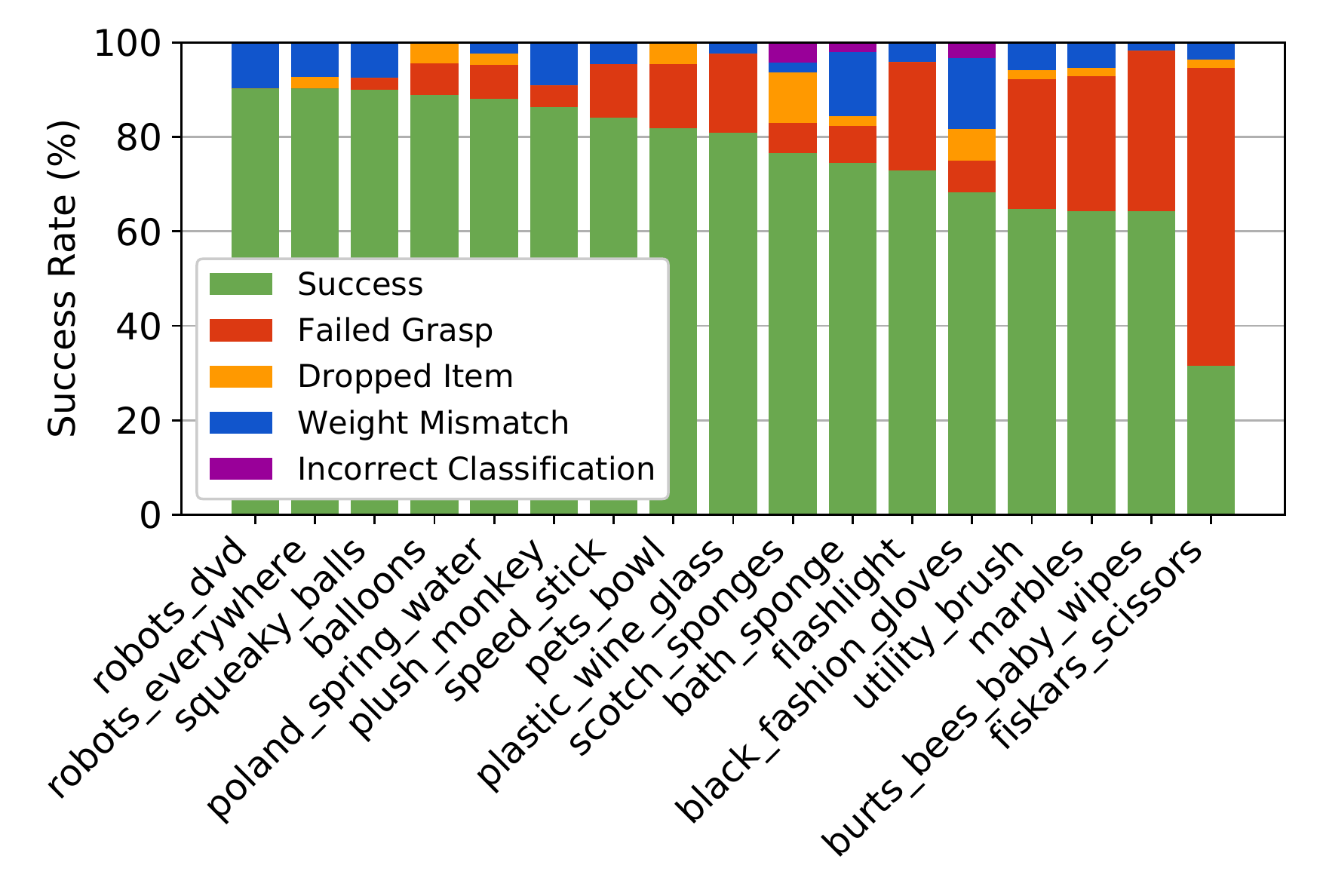}
  \vspace{-7mm}
  \caption{Grasp success rates for the 17 items using during long-term testing.}
  \vspace{-4.5mm}
  \label{fig:successrates}
\end{figure}

The overall grasping success rates per item are shown in Fig.~\ref{fig:successrates}.  Grasp attempt failures were classified as \textit{failed grasp}, where the item was not successfully sucked or gripped, \textit{dropped item}, where the object was successfully sucked or gripped but then dropped before reaching its destination, \textit{weight mismatch}, where an item was grasped but its weight didn't match that of the target object, or \textit{incorrect reclassification}, where an object was successfully grasped but was incorrectly reclassified as a different item based on its weight.

The 168 \textit{failed grasp} attempts can be further categorised by their primary cause, either \textit{perception}, where the failure was caused by an incorrect or incomplete segmentation/identification of the object (27.6\%), \textit{physical occlusion}, where the object was physically occluded by another, resulting in a failed grasp attempt (10.3\%), \textit{unreachable} if the object was in a pose which was physically unobtainable by our gripper such as a small object resting in the corner of the storage system or tote (7.5\%), or \textit{grasp pose failure} if the object was correctly identified and physically obtainable and the grasp failed anyway (54.6\%).  40\% of all failed grasps were on the challenging \texttt{fiskars\_scissors} item, indicating that our manipulator or grasp point detection could be improved for this item.

Table~\ref{tab:speed_results} compares \textit{Cartman}'s performance in the finals task to the other teams' systems, recorded by matching video footage of the competition with score-sheets.  Due to the wide range of strategies used by different teams and the complex environment of the challenge, it is difficult to directly compare the performance of different systems as a whole in the context of the challenge except by points.  However, to highlight the strengths and weaknesses of different systems, we show three metrics which are indicative of different aspects of system performance:
\begin{itemize}
    \item \textit{Grasp Success Rate}: The success rate of the system in executing a grasp attempt, regardless of the item or the action performed with it.  A successful grasp is one where the system lifts an item.  This includes grasping an item to stow/pick it, for item classification or relocating items.
    \item \textit{Average Time per Attempt}: The average time for the system to complete a grasp attempt, i.e. from finishing an action with one item to finishing an action with the next.  This metric is representative of the total execution time of the system, taking into account perception, motion planning, grasp execution and movement.
    \item \textit{Error Rate}: We define the error rate of the system as the ratio of number of penalties (incorrect item labels, incorrect items picked, dropped/damaged items, etc.) incurred to the total number of items stowed and picked during the round, which is indicative of the overall system accuracy.
\end{itemize}

The finals task was chosen as a benchmark for comparison as it represents the final state of the teams' robots and allows for any changes and optimisations made during the competition.  While all teams received a different set of items, similar object classes were chosen by the Amazon team to be of equal difficulty.

\begin{figure}[tpb]
  \centering
  \includegraphics[width=\columnwidth,trim={0.5cm 0.5cm 0.4cm 0},clip]{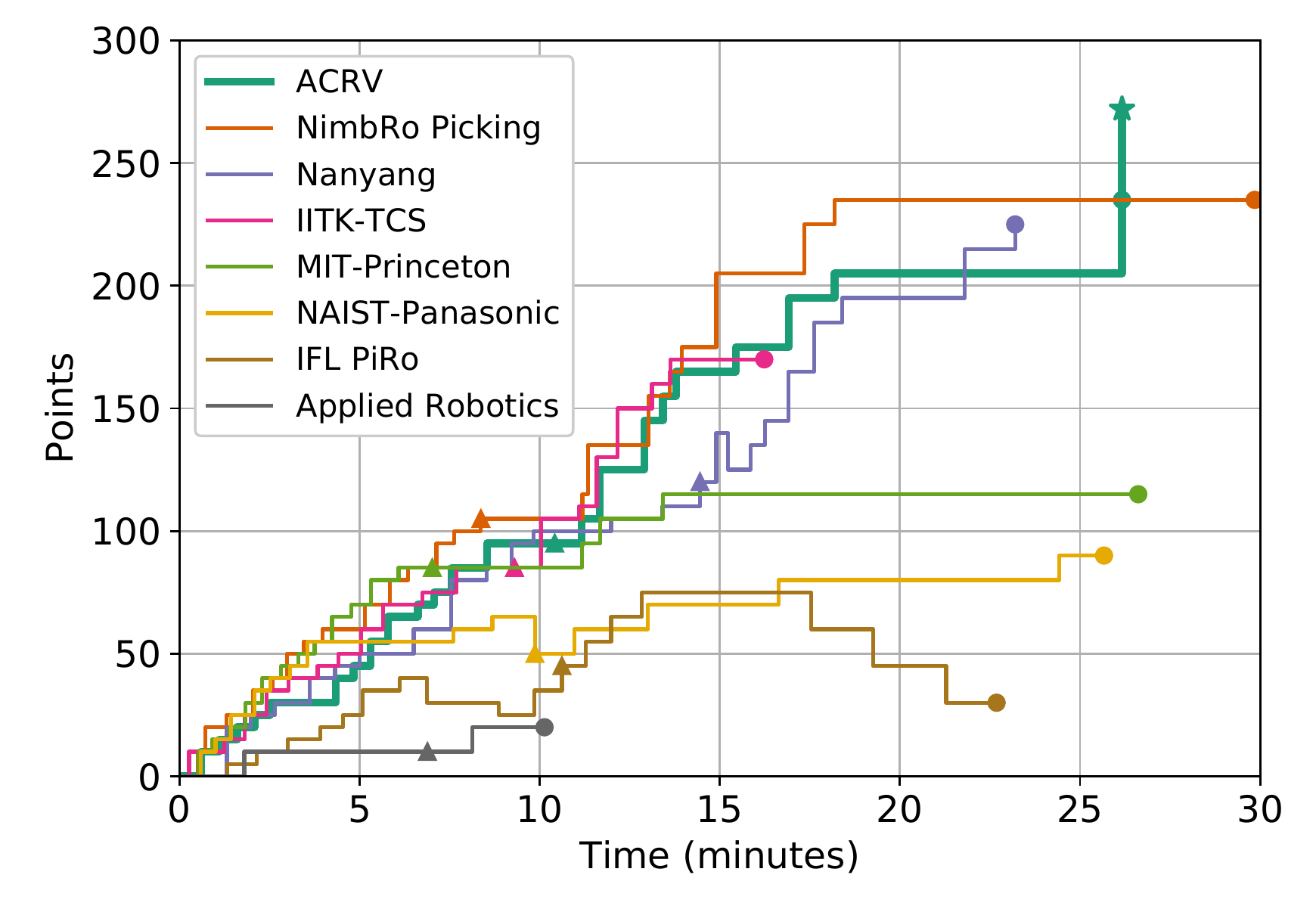}
  \caption{Points accumulated by each team throughout their finals run, recorded by matching video footage of the competition with score-sheets.   Triangles indicate the transition from stow to pick and circles indicate the end of a run.  Stars indicate bonus points for completing the pick task with time remaining.}
  \label{fig:pointsbytime}
  \vspace{-5mm}
\end{figure}

While having a high grasp success rate, low execution time and low error rate are all desirable aspects of an autonomous robot, Table~\ref{tab:speed_results} shows that no one metric is a good indicator of success.  Fig.~\ref{fig:pointsbytime} shows the points accumulated by each team throughout the finals task of the competition, including any penalties incurred, and highlights some of the key differentiating aspects of the teams.  Performance is most consistent between teams during the stow task, with the top five teams stowing roughly the same number of items at a similar, constant rate.  The main separation in points is due to the pick task.  For each team, the rate of acquiring points decreases throughout the pick task, as the difficulty of remaining items and the chances of items being occluded increase, causing many teams to abort their attempt early.  It's here that we attribute our design approach and our system's overall robustness our win.  During the finals round, our system relied on item reclassification, misclassification detection, failed grasp detection, and ultimately our active perception approach to uncover the final item in the pick task, making us the only team to complete the pick task by picking all available ordered items.

\section{Conclusion}

In this paper we presented \textit{Cartman}, our winning entry for the Amazon Robotics Challenge.  The cost-effective, robust Cartesian manipulator was designed and built from scratch in the months leading up to the challenge in July 2017.

The key components of our robotic vision system described in this paper are:
\begin{itemize}
    \item A 6-DoF Cartesian manipulator, featuring independent sucker and gripper end-effectors.
    \item A semantic segmentation perception system capable of learning to identify new items with few example images and little training time.
    \item A multi-level grasp synthesis system capable of working under varying visual conditions.
\end{itemize}

A crucial element of our success is the robust integration of these key components. We describe how our design process helped effectively iterate through a large number of designs. In addition we present lessons-learned from taking part in the challenge two years in a row.

While the performance of robotic systems, as seen during the challenge, is quite a long way from human performance (approximately 400 picks/hour~\cite{correll2016analysis} while \textit{Cartman} can perform about 120 picks/hour), we believe that the following two attributes are critical to winning the challenge and more generally for designing autonomous robotic systems capable of operating in the real world:

\begin{itemize}
    \item A design methodology focused on system-level integration and testing to help optimise competition performance.
    \item High-level logic designed to be robust, and able deal with errors.
\end{itemize}

\bibliographystyle{IEEEtran}
\bibliography{bib}

\end{document}